# Challenges and Opportunities of Evolutionary Robotics

D. A. Sofge, M. A. Potter, M. D. Bugajska, A. C. Schultz
Navy Center for Applied Research in Artificial Intelligence
Naval Research Laboratory
Washington, D.C. 20375, USA
{sofge, mpotter, magda, schultz}@aic.nrl.navy.mil

## Abstract

*Robotic hardware designs are becoming more complex as the variety and number of on-board sensors increase and as greater computational power is provided in ever-smaller packages on-board robots. These advances in hardware, however, do not automatically translate into better software for controlling complex robots. Evolutionary techniques hold the potential to solve many difficult problems in robotics which defy simple conventional approaches, but present many challenges as well. Numerous disciplines including artificial life, cognitive science and neural networks, rule-based systems, behavior-based control, genetic algorithms and other forms of evolutionary computation have contributed to shaping the current state of evolutionary robotics. This paper provides an overview of developments in the emerging field of evolutionary robotics, and discusses some of the opportunities and challenges which currently face practitioners in the field.*

## 1. Introduction

The field of evolutionary robotics has emerged in recent years as the application of artificial evolution to the development of robotic systems. While most of the work in evolutionary robotics has focused on the development of control systems for autonomous mobile robots, some researchers have used the techniques to evolve robotic hardware configurations and even robot body parts along with the controllers.

Evolutionary robotics (ER) has its origins in several disciplines including artificial life [1], cognitive science and neural networks [2,3,4], behavior-based control [5], genetic programming [6] and genetic algorithms [7]. Current practitioners of ER incorporate techniques from a variety of disciplines to achieve the desired result, often a robotic control system for an autonomous mobile robot (or other autonomous system) which exhibits a set of desired behaviors, and for which those behaviors were acquired "automatically" (i.e. without custom programming of each individual behavior).

The need for ER arises from the fact that as robotic systems and the environments into which they are placed increase in complexity, the difficulty of programming their control systems to respond appropriately increases to the point where it becomes impracticable to foresee every possible state of the robot in its environment. Evolutionary algorithms are used to generate control algorithms using the Darwinian principle of survival-of-the-fittest. A population of controllers is maintained and evolved by evaluating individual control systems based on a measure of how well they achieve desired characteristics such as executing appropriate behaviors at appropriate times. Only the fitter members of the population survive and pass the characteristics that made them successful on to future generations. The ultimate goal is to produce the best possible controller given some design criteria. As discussed in the following sections, this approach has proven very successful in a wide variety of challenging robotics domains.

The remainder of this paper will provide a sampling of recent development efforts in evolutionary robotics research with the goal of showing both the opportunities presented by the use of evolutionary techniques for solving difficult problems in robotics, but also in showing some of the challenges of applying these techniques.

## 2. Evolved Controllers for Autonomous Mobile Robots

A key objective in evolutionary robotics is to evolve behavior-based controllers for autonomous mobile robots [8,9]. Autonomous mobile robots often incorporate both reactive and longer-term planning components in order to accommodate goal-driven behaviors. The reactive portion of the controller may be encoded in a variety of forms. Common choices include stimulus-response rules, neural networks, and state-machines. The planning stage may be represented as a series of goal states. Behavior-based controllers are thus driven by a combination of current state, as determined by current sensor readings and possibly short-term memory, and goal(s). The controller attempts to match its current state readings and goal, which together comprise the stimulus part, and then to produce the appropriate output, or response.

## 2.1. Evolved Rule-Based Control

Work by Grefenstette and Schultz [10] resulted in the application of the SAMUEL learning system to evolve stimulus-response rules to produce a reactive control system for autonomous mobile robots. Behaviors achieved using this system include tracking, navigation, and obstacle avoidance. SAMUEL maintains a population of candidate behaviors which are evaluated in a simulated robotic environment. The population is scored, mutation and crossover operators applied, and the population is adjusted to remove the lesser scoring rule sets. The process is run for a number of generations until a stopping criteria is met, at which point the best evolved controller is uploaded to the robot hardware for validation and testing. This technique proved successful for evolving reactive controllers for autonomous mobile robots. Schultz et al. [12] demonstrated the evolution of rule-based controllers for learning complex robotic behaviors by evolving the behavior for a shepherd robot to coerce a sheep robot into a corral. This technique and was extended to evolving controllers for simulated autonomous aircraft and autonomous underwater vehicles [7,11].

Challenges with evolving rule-based controllers include determining how to map continuous inputs and outputs to discrete state variables, establishing appropriate intermediate and goal states, determining how many production rules are required, and performing conflict resolution.

## 2.2. Evolved Neural Network Based Control

An alternative representation used by many researchers in evolutionary robotics is artificial neural networks, which have a number of characteristics that are desirable from an evolutionary robotics perspective. Neural networks are relatively insensitive to noise in the environment since the output of each node is typically a function of the input from a variety of sources. This characteristic also produces a smooth search space with a well-behaved mapping between changes to the network and changes in the resultant network behavior. Neural networks also naturally handle continuous input and can produce either continuous or discrete output as desired [9].

A key advantage of using evolutionary algorithms for producing robotic neural network based controllers is that the evolutionary techniques may be used equally well with feed-forward or recurrent networks [13]. Recurrent networks offer many advantages for dynamic control systems because they allow recent state information to be combined with current state information in the decision-making process. In effect, the recurrent connections provide a kind of short-term memory capability for the neural network so that decisions may include not only the input state information but also the prior state(s) of the network itself. Recurrent networks, however, are notoriously difficult to train using standard gradient-descent learning techniques [14]. Evolutionary algorithms have proven quite successful in training recurrent neural network-based controllers for autonomous mobile robots [9].

Potter et al. [15] demonstrated the evolution of neural network controllers for multiple robots engaged in shepherding a sheep robot. The goal of this work was to examine the effects of evolving a single homogeneous controller for a group of autonomous mobile robots performing a collective herding task, versus coevolving separate heterogeneous controllers, and to determine if the complexity of the task favored homogeneous or heterogeneous control. The heterogeneous controllers were produced using a cooperative coevolutionary architecture [16] in which the controller for each robot is evolved in a separate genetically-isolated species. It was found that heterogeneous controllers are indeed advantageous when the task can be decomposed into subtasks that can be solved by robots specializing in substantially different skill sets. Otherwise homogeneous controllers have an advantage due to their generality.

Quinn et al. [17] described another experiment in which neural network controllers were evolved for a team of real robots. The objective was to study the capability of the robots to learn formation forming behaviors starting from random positions and using a minimal sensor set consisting of 4 IR sensors on each robot.

## 3. Hyper-Redundant Robot Control

One of the greatest challenges in robotics is the design of control systems for robots with high degrees of freedom, particularly if many of the degrees of freedom are coupled. A key example of this is a highly segmented serpentine or snake-like robotic arm (Fig. 1). This is an example of a hyper-redundant robot, where there are many possible kinematic solutions to achieve the same end-effector position or trajectory.

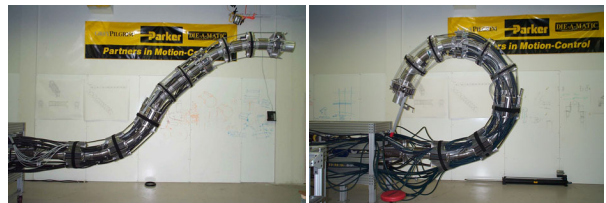

**Figure 1. Hyper-redundant robotic manipulator**

Sofge [18] used evolutionary algorithms to generate the inverse kinematics for the hyper-redundant robotic manipulator. The continuous work space of the robot was discretized into a grid of regularly spaced points. A population of genomes was created such that each genome represented the joint-space configuration of the robot as a real-valued vector. The system then used an evolutionary strategy to coevolve solutions for each grid position within the workspace such that the distance in joint space between any two neighboring points was minimized. The fitness function which was minimized included the joint space distance between neighboring configurations, as well as penalty functions for nearing joint angle limits, exceeding the boundaries of the work space, or putting the manipulator into otherwise undesirable positions.

The resulting coevolved kinematics achieved a position error of less than one inch over a simulated 50 ft. manipulator while maintaining smooth end-point trajectory control from any point to any other point within the work space.

## 4. Evolvable Hardware

Evolvable Hardware (EHW) is an emerging field that applies evolution to automate the design of physically reconfigurable structures such as electronic systems, micro-electromechanical systems (MEMS), and robots. Since EHW techniques enable self-reconfigurability and adaptability of the systems with embedded programmable devices, they have potential to significantly increase the functionality, robustness, and reliability of deployable hardware systems. The benefits of self-reconfiguration and adaptation are evident for space exploration, defense, and other applications that need systems to perform with optimal functionality for extended periods of time in unknown, hostile, and/or changing environments. In autonomous robotics this is especially true.

Evolutionary algorithms have been applied to the design and post-fabrication adaptation of reconfigurable hardware in order to improve the design or performance. The new generation of reconfigurable hardware will also enable the continuous and embedded evolution of robot controllers [19] and vehicle or sensor morphology.

## 5. Coevolution of Robot Bodies and Brains

In nature, both the morphology and the behavior of an organism are evolved in lockstep. In an effort to explore this more natural form of evolution in ER, a number of researchers have begun work coevolving robot hardware designs with software for controlling the robot. Bugajska and Schultz [20,21] demonstrate the coevolution of both rule-based and neural network-based controllers and sensor configurations for autonomous micro-air vehicles (MAVs). The task presented for the simulated MAV is to fly through a forested area to reach a target without hitting any obstacles such as trees. Range sensors with adjustable beam width and maximum range settings are placed at various locations on the. The SAMUEL system was used for evolving the rule-based controllers and sensor configurations on the same chromosome, while the cooperative coevolution architecture [16,22] was used for evolving the neural network controllers and sensor configurations in two separate species. Successes were demonstrated for each method, but more research is needed to better understand the benefits of the use of the cooperative coevolution approach over standard evolutionary algorithms, and to determine if the rule-based or neural network-based representation is better for the controller.

Karl Sims [23] demonstrated the evolution of fantastical simulated creatures in an artificial reality environment by coevolving the creatures' minds and bodies. The creatures' bodies consisted of 3D rectangular blocks, while their minds were evolved neural network controllers. The fitness functions were crafted such that the members of the population that could successfully move across the landscape or track a moving target were more likely to survive and reproduce. After coevolution the creatures exhibited a variety of interesting behaviors resembling walking, hopping, slithering and swimming.

Hornby and Pollack [24,25] studied the coevolution of bodies and brains in simulated robots using L-systems as a generative encoding mechanism. The fitness objective was, as with the Sims work, to evolve for locomotion across a simulated landscape. The result of this effort was to demonstrate that the evolved L-system individuals moved faster than those using standard representations, expressed more modularity of design and exhibited a far greater complexity than their simpler brethren.

In the GOLEM project (Genetically Organized Lifelike Electro Mechanics) Lipson and Pollack [26] coevolved simple neural network controllers and robot body parts as a first step towards creating artificial life-forms. The goal of evolution was to learn locomotion behavior. The creatures were evolved in simulation, and the body parts were automatically constructed using thermoplastic fabrication equipment. The evolved neural network controller was then uploaded to the artificial neural network hardware. Motors, controller, and body parts were hand-assembled by a human assistant. When the creatures were connected to a power source they demonstrated a crude ability to move across a flat surface such as a table or stage.

## 6. Embodied Evolution

Embodied Evolution (EE) is based upon the notion that a group of autonomous robots or agents is situated within an environment trying to perform a task. These robots are allowed to interact with one another, mate, and reproduce control programs which then are transferred to other members of the population [27,28]. The key aspects of this paradigm are that the interactions such as mating, cooperation, and so on are based upon contact between the actual robots in the environment, and the entire population is evaluated in parallel. This inherent parallelism in EE is a major advantage over the more traditional evolutionary approach because as the population size increases, the computational requirements for each robot do not. Thus the evolutionary computation scales well, which is particularly important if it is to be performed on real robots in real time.

Watson et al. [28] describe an early experiment with EE to produce a phototaxis (light-seeking) behavior. Each mobile robot was equipped with two light sensors, two motor control outputs, and infrared diodes to provide omni-directional communications capabilities. The control architecture was a simple neural network model, with inputs from each sensor and outputs to the motors. The evolutionary algorithm used was an adaptation of a steady-state EA model referred to as the probabilistic gene transfer algorithm.

Some of the challenges of getting this system to work were in providing a sufficient number of mobile robots to overcome the stochastic effects of a small population, providing a consistent source of power to the mobile robots, and providing an on-board fitness measure for each robot so that the evolution can be truly distributed. The experiment resulted in a number of solutions comparable to hand-designed behaviors, and a novel looping behavior for reaching the light.

## 7. Simulation vs. Reality

The most natural and direct application of evolutionary computation to robotics is to perform control-system evaluations on real robot hardware in the actual task environment. However, if evolution is done in this way a number of issues become problematic. Evolution is a relatively long time-scale process that may require many control-system evaluations to achieve satisfactory results, leading to unacceptable runtimes. This problem is exacerbated by unreliable hardware and poor battery performance. The robot may also enter into dangerous states in which its hardware could be damaged or bystanders injured, especially in the early stages of evolution. These issues have led most researchers to evolve control systems in simulation where fitness can be evaluated at faster than real-time speed in a safe off-line environment.

Evolution in simulation is not without its own problems. A control system that works well in simulation may not perform satisfactorily on a real robot with noisy sensor readings and imprecise motor control while operating a complex environment that is computationally impractical to model with a high degree of fidelity. Grefenstette et al. [29], in an early attempt to overcome the limitations of evolving control systems in simulation, experimented with adding noise to the sensor models. It was found that given the inevitable mismatch between simulation and reality, it was better to have too much rather than too little noise in the simulation because this encourages more general solutions to emerge. However, a later study by Jakobi et al. [30] showed that unreasonably high levels of noise can also be harmful because solutions may evolve that rely on the excessive noise. By carefully designing a simulation that differentiates between environmental features that are critical to the task at hand and those that are irrelevant, the noise can be tailored as appropriate, thus creating control solutions that ignore the irrelevant features and are robust with respect to the critical ones [31].

Another approach explored by Grefenstette and Ramsey [32] is to combine simulation and reality. In their architecture for anytime learning (also called continuous and embedded learning [33]) control systems are evolved on a simulator that runs continuously on a real robot. Over time, the parameters of the embedded simulation are adjusted to more closely match reality. As new control rules are evolved that have a high likelihood of improving the robot's performance, the active robot control system is updated. This process enables the robot to adapt to mismatches between simulation and reality due to modeling errors, a dynamically changing environment, and changes in the performance characteristics of the robot's sensors and effectors.

## 8. Conclusions

The field of evolutionary robotics is growing rapidly. Robots themselves are quickly becoming more complex as the variety and number of on-board sensors increases, and as Moore's Law results in greater on-board computational power available in ever-smaller packages. Advances in hardware, however, do not automatically translate into better software for controlling complex robots. With an increased emphasis on more capable and more autonomous robots, evolutionary techniques hold the potential to solve many difficult problems in robotics which defy simple conventional approaches.

Behaviorism has become the predominant paradigm for control of autonomous robots. Artificial evolution provides a means to achieve desired behaviors. Evolutionary techniques also hold the promise of having the robots learn new behaviors automatically in order to adjust to changes in their environments, changes in themselves due to sensor drift or malfunction for example, and to acquire skills for performing new tasks or improving on old tasks.

As we develop ever more capable robots which assist humans in performing tasks, it is becoming clear that we must also improve the interfaces and means of interacting with these robots. One approach we are pursuing at the Naval Research Laboratory is known as *embodied cognition* [34]. In embodied cognition we develop cognitive models of human performance to augment a robot's reasoning capabilities. Embodied cognition is based on the premise that use of cognitive models on the robot facilitates human-robot interaction by making it easier for humans to predict and understand the robot's behavior and to interact with the robot. This embodied cognition will sit on top of evolved reactive behaviors and will allow future systems to have higher-level cognitive abilities.

For ER to achieve the vision of supplanting the hand-coding and hand-design of complex robotic systems, as pointed out by Mataric [35] it must be shown to reduce the overall effort of its programmers and designers. The evolved controllers produced by current ER systems are often very simple and could be easily bettered by duly considered hand-coded solutions. Representations for robot morphologies and controllers do not often exploit notions of modularity which we expect in more complex creatures. Better representations and better evolutionary algorithms may be necessary to achieve the vision of ER, and better techniques for transferring these solutions into robot hardware, or evolving them *in situ*, are needed.

## Acknowledgments

This work was supported by the Office of Naval Research under work order requests N0001402WX20374, N0001403WR20212 and N0001403WR20057